\documentclass[11pt]{article}
\usepackage{fullpage}
\makeatletter
\long\def\@makecaption#1#2{
  \vskip 0.8ex
  \setbox\@tempboxa\hbox{\small {\bf #1:} #2}
  \parindent 1.5em 
  \dimen0=\hsize
  \advance\dimen0 by -3em
  \ifdim \wd\@tempboxa >\dimen0
  \hbox to \hsize{
    \parindent 0em
    \hfil 
    \parbox{\dimen0}{\def\baselinestretch{0.96}\small
      {\bf #1.} #2
     
    } 
    \hfil}
  \else \hbox to \hsize{\hfil \box\@tempboxa \hfil}
  \fi
}
\makeatother

\usepackage{times,enumitem,hyperref,booktabs,multicol,blindtext,url,parskip}

\usepackage{pifont}

\usepackage{adjustbox}
\usepackage{amsfonts}
\usepackage{amsmath}
\usepackage{amsthm}
\usepackage{amssymb}
\usepackage{microtype}
\usepackage{graphicx}
\usepackage{natbib}
\usepackage{fancyhdr}
\usepackage{color}
\usepackage{algorithm}
\usepackage{forloop}
\usepackage{amsthm}
\usepackage[english]{babel}
\usepackage[T1]{fontenc}
\usepackage{amsmath}
\usepackage{amssymb}
\usepackage{amsfonts}
\usepackage{multirow}
\usepackage{mathtools}
\usepackage{breqn}
\usepackage{bbm}
\usepackage{dsfont}
\usepackage{graphicx}
\usepackage{natbib}
\usepackage{cases}
\usepackage[colorinlistoftodos]{todonotes}

\renewcommand{\hat}{\widehat}

\newtheorem*{remark*}{Remark}

\newtheorem*{observation*}{Observation}

\numberwithin{equation}{section}

\newcommand{\Gnorm}[1]{{\left\vert\kern-0.25ex\left\vert\kern-0.25ex\left\vert #1 
		\right\vert\kern-0.25ex\right\vert\kern-0.25ex\right\vert}}
\newcommand{\gnorm}[1]{{\vert\kern-0.25ex\vert\kern-0.25ex\vert #1 
		\vert\kern-0.25ex\vert\kern-0.25ex\vert}}

\usepackage{multirow}
\usepackage{enumerate}
\usepackage{algpseudocode}
\usepackage{rotating}
\usepackage{makecell}
\usepackage{bm}
\usepackage{wrapfig}

\usepackage[capitalize]{cleveref}
\crefname{section}{Sec.}{Secs.}
\Crefname{section}{Section}{Sections}
\Crefname{table}{Table}{Tables}
\crefname{table}{Tab.}{Tabs.}

\def\shownotes{0}  
\ifnum\shownotes=1
\newcommand{\authnote}[2]{[#1: #2]}
\else
\newcommand{\authnote}[2]{}
\fi

\definecolor{mypink}{RGB}{219, 48, 122}

\usepackage{xcolor}

\hypersetup{
  colorlinks,
  citecolor=violet,
  linkcolor=red,
  urlcolor=blue}

\begin{document}

\abovedisplayskip=8pt plus0pt minus3pt
\belowdisplayskip=8pt plus0pt minus3pt

\begin{center}
  \vspace*{-1cm}
  {\LARGE Towards Micro-Action Recognition with Limited Annotations: \\ An Asynchronous Pseudo Labeling and Training Approach
  \vspace{0.12cm}
  ~} \\
  \vspace{.4cm}
  {\Large Yan Zhang,~Lechao Cheng,~Yaxiong Wang,~Zhun Zhong,~Meng Wang
  } \\
  \vspace{.4cm}
  Hefei University of Technology, Hefei, China

\end{center}

\begin{abstract}
Micro-Action Recognition (MAR) aims to classify subtle human actions in video. However, annotating MAR datasets is particularly challenging due to the subtlety of actions. To this end, we introduce the setting of Semi-Supervised MAR (SSMAR), where only a part of samples are labeled. We first evaluate traditional Semi-Supervised Learning (SSL) methods to SSMAR and find that these methods tend to overfit on inaccurate pseudo-labels, leading to error accumulation and degraded performance. This issue primarily arises from the common practice of directly using the predictions of classifier as pseudo-labels to train the model. 
To solve this issue, we propose a novel framework, called Asynchronous Pseudo Labeling and Training (APLT), which explicitly separates the pseudo-labeling process from model training. 
Specifically, we introduce a semi-supervised clustering method during the offline pseudo-labeling phase to generate more accurate pseudo-labels. Moreover, a self-adaptive thresholding strategy is proposed to dynamically filter noisy labels of different classes. We then build a memory-based prototype classifier based on the filtered pseudo-labels, which is fixed and used to guide the subsequent model training phase. By alternating the two pseudo-labeling and model training phases in an asynchronous manner, the model can not only be learned with more accurate pseudo-labels but also avoid the overfitting issue.
Experiments on three MAR datasets show that our APLT largely outperforms state-of-the-art SSL methods. For instance, APLT improves accuracy by 14.5\% over FixMatch on the MA-12 dataset when using only 50\% labeled data. Code will be publicly available.
\end{abstract}

\section{Introduction}
\begin{figure}[!t]
\centering
\includegraphics[width=.75\linewidth]{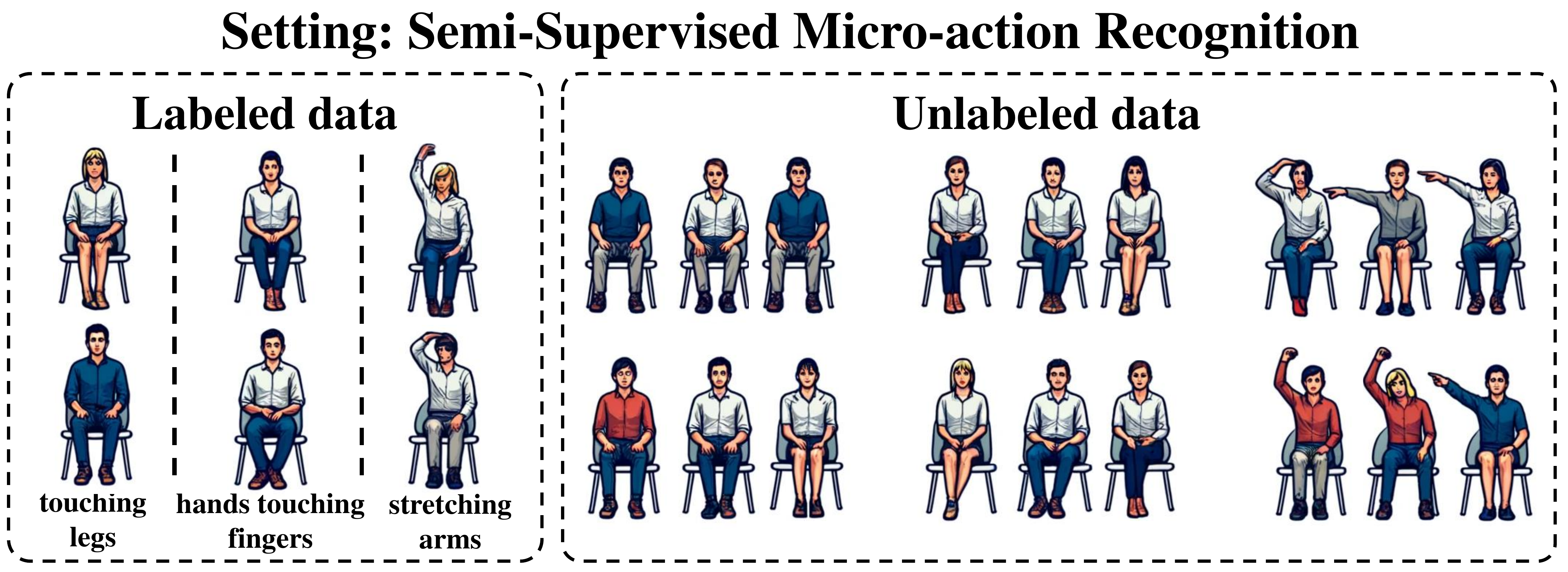}
\caption{We present a new setting, Semi-Supervised Micro-Action Recognition (SSMAR), which aims to train a model that can recognize subtle, rapid micro-actions in videos by utilizing both labeled and unlabeled data.}
\label{fig:setting}
\end{figure}

Micro-action refers to the fast and tiny movements or body language signals that exhibit by humans during communication~\cite{Noroozi_2021}, which plays a significant role in revealing individuals' inner states and true intentions.
Traditional action recognition~\cite{arnab2021vivit,feichtenhofer2019slowfast,xie2018rethinking} typically focuses on identifying and classifying overt movements, \textit{e.g.}, running, jumping, etc. In contrast, Micro-Action Recognition (MAR)~\cite{chen2023smg,liu2021imigue} targets the detection and analysis of minor movement changes, which are imperceptible to human eyes, making MAR inherently more challenging. 
Existing MAR methods are mainly developed under  a fully-supervised context, which heavily depend on large amounts of high-quality labeled data~\cite{caba2015activitynet,kay2017kinetics,soomro2012ucf101}. 
However, acquiring labeled MAR datasets is labor-intensive and costly due to the subtlety of micro-action and privacy concerns. To alleviate the challenges of labeling MAR datasets, we introduce a Semi-Supervised Learning (SSL) setting for MAR, called Semi-Supervised MAR (SSMAR). As shown in Fig.~\ref{fig:setting}, SSMAR aims to harness both labeled and unlabeled data to train a MAR model, enabling performance close to that of a model trained on fully-labeled data.

\begin{figure*}[!t]
\centering
\includegraphics[width=.95\linewidth]{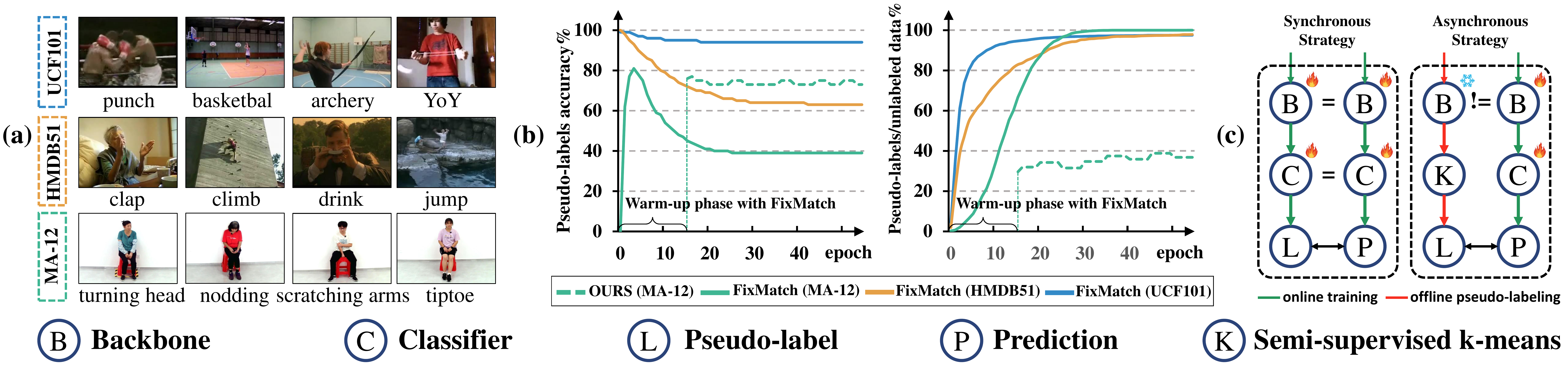}
\caption{(a) Micro-actions (MA-12) are less distinct from each other and more difficult to differentiate than conventional actions (UCF101 and HMDB51). (b) Comparison of training process between our method and Fixmatch on the traditional action recognition datasets and a MAR dataset with 50\% labeled data. FixMatch performs well on traditional action recognition datasets (HMDB51 and UCF101). However, when applying it on a MAR dataset (MA-12), as training proceeds, the number of unlabeled samples that pass the set threshold for the online pseudo-labeling method gradually increases, and the accuracy of the pseudo-labeling gradually decreases. In contrast, our method consistently generates high-accurate pseudo-labels. (c) 
FixMatch performs a synchronous pseudo-labeling and training. Instead, the proposed asynchronous approach separates the pseudo-labeling from the training process, where the pseudo-labels are first obtained by semi-supervised clustering in the offline phase and are then utlized for the online model training.} 
\label{fig:introduction}
\end{figure*}

SSL has been well established in many fields, such as image classification~\cite{sohn2020fixmatch,berthelot2019mixmatch}, semantic segmentation~\cite{chen2021semi,luo2020semi}, action recognition~\cite{xing2023svformer,wu2023neighbor}, etc. The popular SSL methods mainly focus on consistent regularization~\cite{Rasmus_Valpola_Honkala_Berglund_Raiko_2015,Ke_Wang_Yan_Ren_Lau_2019} and pseudo-labeling strategies~\cite{lee2013pseudo,xie2020self}  (see Fig.~\ref{fig:introduction}, (c) left). However, directly applying these methods to the SSMAR task presents a new challenge, \textit{i.e.}, the model tends to accumulate pseudo-labeling errors more significantly in the later stages of training. As shown in Fig.~\ref{fig:introduction} (b), on a traditional action recognition dataset,  UCF101~\cite{soomro2012ucf101}, the popular method FixMatch~\cite{sohn2020fixmatch} is effective in mining a large number of accurate pseudo-labels. However, as the complexity of the action recognition task increases, such as HMDB51~\cite{kuehne2011hmdb}, the accuracy of pseudo-labels declines significantly and the issue of overfitting to incorrect pseudo-labels becomes increasingly severe. The problem is particularly acute in the context of MAR dataset, \textit{e.g.}, MA-12~\cite{guo2024benchmarking} (examples of different tasks are shown in Fig.~\ref{fig:introduction} (a)). Specifically, when applying FixMatch to the MA-12 dataset, the accuracy of pseudo-labels drops sharply from an initially high level (80\%) after early training stages. Meanwhile, the model grows increasingly confident in its predictions and assigns pseudo-labels to all samples. These two observations indicate that FixMatch leads to an increasing number of incorrect pseudo-labels in late training stages. We argue that the primary cause of the overfitting issue is that the model relies on its own predictions as pseudo-labels while simultaneously using those same pseudo-labels as supervisory signals for training. This approach works well when the task is relatively simple, as shown with UCF101, where it generates a large number of accurate pseudo-labels. However, as task complexity grows, this approach becomes more prone to inaccurate pseudo-labels, leading to error accumulation and performance degradation, as observed on the MAR dataset.

To tackle this challenge, we introduce a novel framework, called Asynchronous Pseudo Labeling and Training (APLT), which explicitly decouples the pseudo-labeling from the model training process (see Fig.~\ref{fig:introduction}, (c) right).
The framework operates in two phases: 1) offline pseudo-labeling that aims to generate accurate pseudo-labels through a semi-supervised clustering approach, and 2) online training that focuses on training the model using a prototype classifier in a robust manner. 
\textit{In the offline pseudo-labeling phase}, we propose a non-parametric, semi-supervised k-means clustering algorithm to generate accurate pseudo-labels. 
Specifically, we use labeled data as anchor points to cluster unlabeled data, thereby generating initial pseudo-labels. To enhance stability, we introduce a labeled-augmentation technique that increases the diversity of labeled samples, making the clustering anchors more robust for guiding the clustering process.
In addition, a self-adaptive thresholding strategy is proposed to filter out less reliable pseudo-labels.
We then build a memory-based prototype classifier by averaging the features of samples that are assigned with pseudo-labels for each cluster. 
\textit{In the online training phase}, the pseudo-labels generated in the offline phase are used to supervise the outputs of the memory-based prototype classifier. 
The prototype classifier remains fixed during training and is updated only in the offline phase. The offline and online phases are performed alternately, helping the model to avoid the overfitting problem in an asynchronous manner. 
Moreover, we reduce the update frequency of pseudo-labeling to several epochs instead of one epoch, further mitigating the overfitting risk. As shown in Fig.~\ref{fig:introduction} (b), our method is more resistant to overfitting on inaccurate pseudo-labels compared to FixMatch.

The main contributions are summarized as follows:
\begin{itemize}

\item We introduce a new setting, SSMAR, designed to reduce the annotation requirements of MAR. Additionally, we identify a critical challenge in applying the FixMatch to SSMAR, \textit{i.e.}, overfitting on incorrect pseudo-labels.

\item We propose a novel framework for SSMAR that explicitly separates the pseudo-labeling process from model training, making them asynchronous. This strategy can effectively mitigate the overfitting issue.

\item Within our framework, we propose several strategies to enhance the reliability of pseudo-labels during the offline pseudo-labeling phase. Furthermore, we develop a memory-based prototype classifier to mitigate the overfitting issue during the online model training phase.

\item  Experiments on three SSMAR benchmarks demonstrate that our method significantly improves the performance of FixMatch, achieving state-of-the-art results.
\end{itemize}

\begin{figure*}
    \centering
    \includegraphics[width=.95\linewidth]{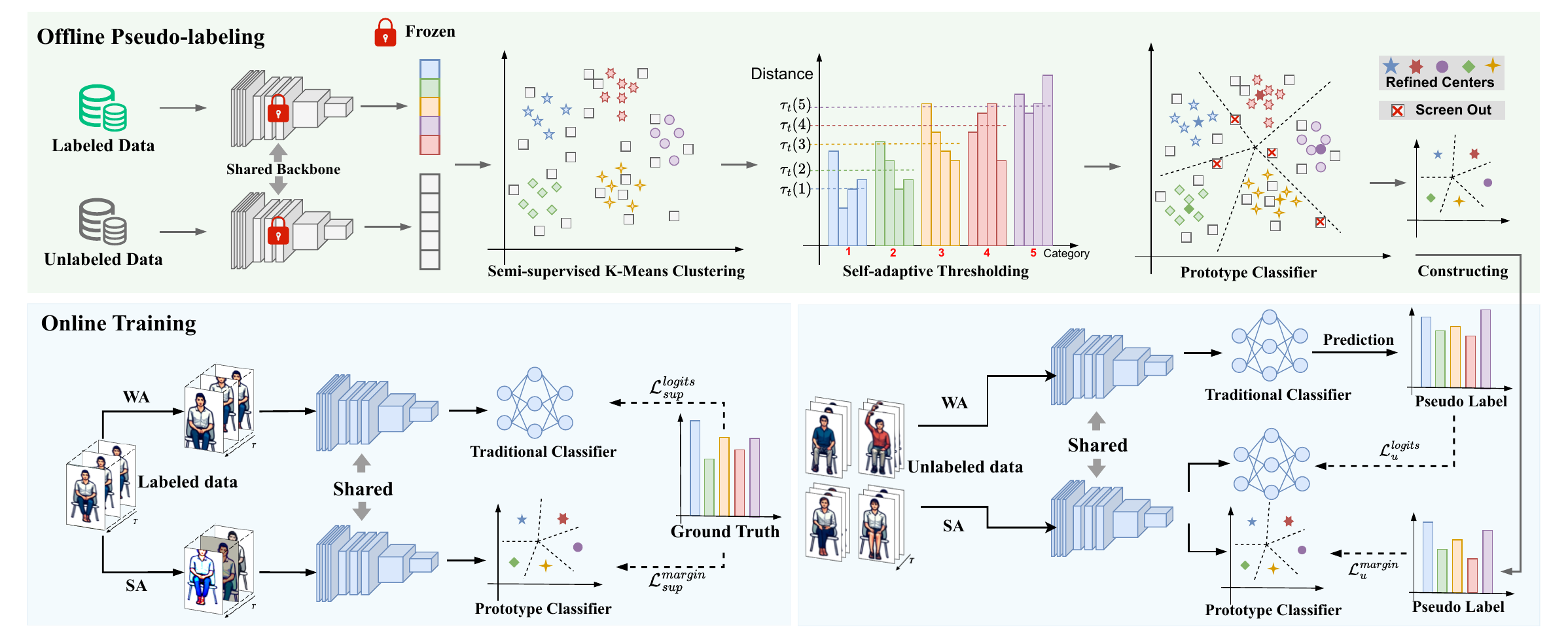}
    
    \caption{Overview of the proposed APLT framework. APLT includes two phases: offline pseudo-labeling and online model training. During the offline phase, we propose an approach to generate reliable pseudo-labels by semi-supervised clustering and self-adaptive thresholding. In addition, we construct a memory-based prototype classifier by averaging features assigned with the same cluster. During the online phase, we augment samples for both labeled and unlabeled samples. For the labeled data, we use the ground-truth labels to supervise the two classifiers ($\mathcal{L}^{margin}_{sup}$ and $\mathcal{L}^{logits}_{sup}$). For the unlabeled data, we use the predictions of traditional classifier to supervise the same classifier ($\mathcal{L}^{logits}_{u}$) while use the pseudo-labels generated by the offline phase to supervise the prototype classifier ($\mathcal{L}^{margin}_{u}$). ``WA'' and ``SA'' stand for weak augmentation and strong augmentation, respectively.
}
    \label{fig:method}
\end{figure*}

\vspace{-.1in}
\section{Related Works}

\noindent\textbf{Micro-Action Recognition (MAR).} Unlike the traditional action recognition, MAR focuses on classifying subtle and transient human body movements. To facilitate the study of MAR, multiple MAR datasets are proposed. The iMiGUE dataset~\cite{liu2021imigue} combines fine-grained gesture actions with sentiment labels for micro-gesture understanding and sentiment analysis. Similarly, Chen et al.~\cite{chen2023smg} propose the SMG dataset, where participants perform various micro-gestures by narrating both false and true stories. Building on this, the MA-52 dataset~\cite{guo2024benchmarking} is proposed by capturing natural micro-actions from psychological interviews. Traditional action recognition~\cite{wang2018temporal,lin2019tsm,bertasius2021space} has been benefited from large-scale labeled datasets~\cite{caba2015activitynet,kay2017kinetics,soomro2012ucf101} to achieve high accuracy. However, obtaining labeled MAR datasets is much more difficult. \textit{The proposed SSMAR setting aims to utilize both labeled and unlabeled data to train a MAR model that can be comparable with the one trained with fully-labeled data.}

\noindent\textbf{Semi-Supervised Learning (SSL).} SSL has gained significant attention in the community, aiming to leverage abundant unlabeled data alongside limited labeled data to enhance model performance. Two prominent SSL trends are pseudo-labeling and consistency regularization. Pseudo-labeling approaches~\cite{lee2013pseudo,xie2020self,pu2023dynamic,pu2024federated} rely on adding high-confidence pseudo-labels to the training dataset. Consistency regularization approaches~\cite{Rasmus_Valpola_Honkala_Berglund_Raiko_2015,Ke_Wang_Yan_Ren_Lau_2019,oliver2018realistic,pu2020dual,pu2021lifelong,pu2023memorizing} assume that applying perturbations to input samples or features does not change the outputs of the model. FixMatch~\cite{sohn2020fixmatch} combines pseudo-labeling and consistency regularization by using high-confidence pseudo-labels of weakly augmented samples to supervise strongly augmented samples. SoftMatch~\cite{chen2023softmatch} uses a Gaussian function to assign weights to samples, resolving the trade-off between quantity and quality of pseudo-labels. \textit{The above methods mainly are designed for image classification task. Instead, we propose a novel SSL framework for the MAR task, which performs pseudo-labeling and model training in an asynchronous way.}

\noindent\textbf{Semi-Supervised Action Recognition (SSAR).} The exploration of SSL in video recognition lags behind the progress in image classification.
VideoSSL~\cite{jing2021videossl} compares SSL methods that are specifically applied to videos, revealing limitations in extending pseudo-labeling directly. 
LTG~\cite{xiao2022learning} introduces temporal gradient as an additional modality to generate high-quality pseudo-labels for training. 
 Recently, self-supervised learning has proven to be effective in learning powerful video representations~\cite{dave2022tclr,Feichtenhofer_Fan_Xiong_Girshick_He_2021,qian2021spatiotemporal}. TimeBalance~\cite{dave2023timebalance}  leverages the temporal contrastive losses from TCLR~\cite{dave2022tclr} to learn the temporal distinctive teacher.
SVFormer~\cite{xing2023svformer} explores the potential benefit of Video Transformers for SSAR. FinePseudo~\cite{dave2025finepseudo} is proposed to improve pseudo-labeling through temporal alignment for fine-grained action recognition under SSL context. \textit{Different from them, this work focuses on the SSMAR task, which is more difficult than traditional and fine-grained action recognition tasks.}

\section{Method}

\noindent \textbf{Task Definition.} In SSMAR, we are given a set of human micro-action videos, defined as $\mathcal{D}=\{\mathcal{D}_{l},\mathcal{D}_{u}\}$, where samples of $\mathcal{D}_{l}$ are labeled while samples of $\mathcal{D}_{u}$ are unlabeled. 
$\mathcal{D}_{l} = \{V^{i}_{l}, y^{i}_{l}\}_{i=1}^{N_{l}}$ consists of $ N_{l}$ videos and corresponding labels.
The videos are from $C$ categories, \textit{i.e.}, the label $y^{i}_{l}$ is derived from the set of labels $Y=\{1,2,\cdots, C\}$.
$\mathcal{D}_u = \{U^{i}\}_{i=1}^{N_u}$ consists of $ N_{u}$ unlabeled videos. 
The goal of SSMAR is to leverage both labeled and unlabeled data to learn an effective MAR model.

\subsection{Overview}
As shown in Fig.~\ref{fig:method}, we propose an asynchronous pseudo-labeling and training framework, dubbed APLT, for SSMAR, which includes two phases: offline pseudo-labeling and online model training.
\textit{During offline phase}, we propose a non-parametric, semi-supervised k-means clustering method to obtain accurate pseudo-labels. In particular, labeled data serve as anchors to guide the clustering, producing initial pseudo-labels for unlabeled data. To improve robustness, we employ a labeled-augmentation approach that enhances labeled sample diversity, making the clustering anchors more dependable for guiding the clustering process. Additionally, a self-adaptive thresholding mechanism is presented to filter out less reliable pseudo-labels. Then, we construct a memory-based prototype classifier by averaging the features of samples assigned with the same cluster.
 \textit{During online phase}, pseudo-labels produced in the offline phase are used to supervise the memory-based prototype classifier’s outputs. This classifier remains unchanged throughout training and is only updated in the offline phase. By alternating between offline and online phases, the model effectively avoids overfitting in an asynchronous manner. Note that, the FixMatch is also applied in our framework in default, which uses the predictions of the parametric classifier as pseudo-labels to supervise the same classifier.

\subsection{Warm-Up}
Our asynchronous learning strategy is implemented after a warm up stage with FixMatch. Specifically, given a labeled video $V^i_l$, we calculate the cross-entropy loss for labeled set by:
\begin{equation}
\mathcal{L}^{logits}_{sup}=\frac1B\sum_{i=1}^B\mathcal{H}(y^i_l, p_m(y\mid\omega(V^i_l))),
\end{equation}
where $B$ is the batch size, $\omega(\cdot)$ indicates the weak data augmentation function, $p_m(\cdot)$ is the output probability from the parametric classifier, $y^i_l$ is the ground-truth label for $V^i_l$, and $\mathcal{H}(\cdot,\cdot)$ is the cross-entropy loss. 

For the unlabeled set, we use a similar loss function but generate pseudo-labels for unlabeled data as we do not have ground-truth labels for them. The loss for unlabeled data is formulated as:
\begin{equation}
\ell^{logits}_u=\frac{1}{B}\sum_{i=1}^{B} \mathbf{1} (\max(q_i)\geq\tau)\mathcal{H}(\hat{q}_i, p_m(y\mid\mathcal{A}(U^{i}))),
\end{equation}
\normalsize
where $q_i$ is the predicted class distribution  of unlabeled video data after weak augmentation: $q_i=p_m(y\mid\omega(U^{i}))$. $\hat{q}_i=\arg\max(q_i)$ is the predicted pseudo-label. $\mathcal{A}(\cdot)$ means the strong data augmentation function. $\mathbf{1}(\cdot)$ is an indicator function of the confidence-based threshold and $\tau$ is the threshold.

Overall, the loss function for warm up stage is:
\begin{equation}
\label{eq:fixmatch}
\mathcal{L}^{logits}=\mathcal{L}^{logits}_{sup} + \mathcal{L}^{logits}_u.
\end{equation}

After training the model with the basic FixMatch for several epochs, we will additionally include the proposed asynchronous method, which includes the offline pseudo-labeling and online model training phases.

\subsection{Phase \text{I}: Offline Pseudo-Labeling}

To mitigate the risk of overfitting on incorrect supervisions during training, we propose to generate pseudo-labels by a non-parametric clustering method instead of directly using the output of the parametric classifier, which includes three strategies: semi-supervised clustering, labeled-augmentation, and self-adaptive threshold strategy.

\noindent\textbf{Semi-supervised Clustering.} The key of semi-supervised clustering is using labeled data as anchors to guide the clustering on unlabeled data, which is inspire by \cite{vaze2022generalized}.
Specifically, after a training epoch, we first generate features for all data, $F= \{F_u, F_{l}\}$, in which the feature is obtained by the feature extractor $ f_\theta$. $F_l$ and $F_{u}$ represent the features of labeled and unlabeled data respectively.
We then initialize the $C$ centroids by averaging the features of labeled data for each class and implement the following two stages.
1) Cluster Assignment: Each unlabeled instance is assigned with a cluster label by identifying its nearest centroid. 2) Center Update: The centroids are updated by averaging all data features within each cluster, where the cluster labels are ground-truth labels for labeled data and are the pseudo-labels generated by first stage (cluster assignment) for unlabeled data, respectively.
We iteratively repeat the above two stages until the algorithm converges, resulting in clustering pseudo-labels for unlabeled data.

\noindent\textbf{Labeled-Augmentation.} To improve the clustering stability, we introduce the labeled-augmentation by applying strong augmentation to the labeled data during the semi-supervised clustering. In this way, we can obtain the augmented labeled set $\mathcal{D}_{sl}$ which is also utilized to initialize and update centers in the clustering process. This strategy enables us obtain more robust clustering centers and thus produce more accurate pseudo-labels for unlabeled data.

\noindent\textbf{Self-Adaptive Threshold Strategy.} 
During clustering, hard samples will be assigned with wrong clustering labels. Using such incorrect pseudo-labels for model training will hamper the optimization. Thus, it is important to select reliable pseudo-labels. One common solution is using a threshold to filter out pseudo-labels with low-confidence. However, since micro-action categories are diverse, the difficulties of recognizing them are very different. Thus, using a single fixed threshold to constrain all classes is not reasonable. For example, the average confidence level for the head movement-related category is typically higher than that for the upper limb movement-related category, as recognizing the former category is more easy.

To solve this problem, we propose a self-adaptive thresholding strategy, which is calculated by both global threshold and category-specific local thresholds. Specifically, given an unlabeled sample, we use its distance from the assigned centroid as the confidence indicator.
The global threshold $\tau_{global}$ is set as the average distance of unlabeled data from their corresponding assigned clusters, reflecting the overall clustering status. 
Local thresholds respond to the clustering state of each class, in which we compute the class-specific local threshold for each cluster. 
By considering both global and local thresholds, our self-adaptive thresholding strategy can be expressed as:
\begin{equation}
\label{eq:ad-thre}
{
\left\{
\begin{aligned}
\tau_{global} &= \frac{1}{N_u}\sum_{i=1}^{N_u}dis(U^{i}),\\
\tau_{local}(c) &= \frac{1}{N^{c}_u}\sum_{i=1}^{N^{c}_u}dis(U^{i}_{c}),\\
\tau_{adapt}(c) &= \frac{\tau_{local}(c)}{\max(\tau_{local})}
\cdot\tau_{global },
\end{aligned}
\right.
}
\end{equation}
where $U^{i}_{c}$ represents an instance assigned with category $c$ in $\mathcal{D}_{u}$ and $N^{c}_u$ is the number of unlabeled instances assigned with category $c$. 
$dis(.)$ indicates the distance to the assigned class center for an unlabeled instance. 
$\tau_{adapt}(c)$ is the adaptive threshold of class $c$.  $\tau_{local}=[\tau_{local}(1), \tau_{local}(2), \ldots, \tau_{local}(C)]$ is the set of local thresholders.
Given an unlabeled sample $U^{i}$, if its distance to the assigned center $c$ is lower than $\tau_{adapt}(c)$, we regard its predicted cluster label as a reliable pseudo label. Otherwise, we ignore its pseudo label. By doing so, we could obtain a filtered pseudo-label set $\mathcal{D}_{su} = \{U^{i}_{p}, y^{i}_{p}\}_{i=1}^{N_{p}}$ with $N_p$ instances.

\noindent\textbf{Memory-based Prototype Classifier.} 
Most of the previous SSL approaches rely on the supervision of the output of the parametric-classifier during training on unlabeled data. 
Instead, we build a non-parametric memory-based prototype classifier based on the features of all data. Specifically, for a class $c$, we average all features of labeled data with class label $c$ and unlabeled  data assigned with pseudo-label $c$,  obtaining corresponding prototype feature:
\begin{equation}
\label{eq:MPC}
\rho_c=\frac1{N_c}\sum_i^{N_c}F_{c}^{i},
\end{equation}
where $N_c$ is the number of samples belonging to class $c$. $F_{c}^{i}$ is the feature of an instance belonging to class $c$. Thus, the prototype feature is $\rho = \{\rho_1,\rho_2,\cdots,\rho_C\}$, constructing the memory-based prototype classifier. This non-parametric classifier is directly added after the backbone for model training, like the parametric one.

\subsection{Phase \text{II}: Online Training}

During the online training, we not only train the model with the basic FixMatch loss based on parametric classifier but also with a margin loss based on the non-parametric classifier. Specifically, given a sample with strong data augmentation, we first calculate its feature and obtain the prediction based on the non-parametric classifier. The margin loss for labeled data is formulated as:
\begin{equation}
\label{eq:loss_margin_labeled}
\mathcal{L}^{margin}_{sup}=-\frac1B\sum_{i=1}^B\log\frac{exp(F^i_l\cdot \rho_{c})}{\sum_{i=1}^Cexp(F^i_l\cdot\rho_i)},
\end{equation}
where $F^i_l$ is the feature of the labeled sample $i$ obtained by the current training model and $\rho_{c}$ is the prototype feature corresponding to the ground-truth $c$.

Similarly, the margin loss for unlabeled data is as:
\begin{equation}
\label{eq:loss_margin_unlabeled}
\mathcal{L}^{margin}_{u}=-\frac1B\sum_{i=1}^B\log\frac{exp(F^i_u\cdot \rho_{c})}{\sum_{i=1}^Cexp(F^i_u\cdot\rho_i)},
\end{equation}
where $F^i_u$ is the feature of the unlabeled sample $i$ and $\rho_{c}$ is the prototype feature of the assigned pseudo-label $c$ of the unlabeled sample $i$.

The final margin loss can be formulated as:
\begin{equation}
\label{eq:loss_margin}
\mathcal{L}^{margin}=\mathcal{L}^{margin}_{sup} +\mathcal{L}^{margin}_{u}.
\end{equation}

By combining with the basic Fixmatch, our final loss is defined as:
\begin{equation}
\label{eq:loss_all}
\mathcal{L}=\mathcal{L}^{logit}+\lambda \mathcal{L}^{margin},
\end{equation}
where $\lambda$ is a hyperparameter that balances the contribution between $\mathcal{L}^{logit}$ and $\mathcal{L}^{margin}$.

 \begin{table*}[!t]
  \caption{Performance (\%) comparison with traditional SSL methods. Results are evaluated on of MA-12, SMG-5 and iMiGUE-11 ($\uparrow$).}
  
  \label{exp_MA-52}
  \centering
  \begin{adjustbox}{width=\textwidth}
  \begin{tabular}{l|cccc|cccc|c|c}
    \toprule
 & \multicolumn{8}{c|}{\textbf{MA-12}} & \textbf{SMG-5}&\textbf{iMiGUE-11}\\
     \cmidrule(r){2-11}
    & \multicolumn{4}{c|}{\textbf{Resnet-18}} & \multicolumn{4}{c|}{\textbf{Resnet-50}} &   \textbf{Resnet-18}&\textbf{Resnet-18}\\
    \cmidrule(r){2-11}
    \textbf{Method}& \textbf{10\% }& \textbf{25\% }& \textbf{40\% }& \textbf{50\% }& \textbf{10\% }& \textbf{25\% }& \textbf{40\% }& \textbf{50\% }& \textbf{20\% }&\textbf{30\% }\\
    \midrule
    Baseline (Labeled Only)& 24.8 & 30.7 & 35.8 & 36.3 & 23.8 & 31.7 & 34.8 & 38.0  & 51.6 
&38.0
\\
\toprule
    Pseudolabel~\cite{lee2013pseudo} & 26.9 & 32.7 & 34.6 & 35.6 & 17.6 & 29.8 & 40.1 & 39.3  & 57.2
&32.3\\
    Mean Teacher~\cite{tarvainen2017mean} & 21.5 & 26.0 & 31.0 & 14.4 & 20.0 & 23.6& 37.5& 34.5& 48.0&30.9\\
    VAT~\cite{miyato2018virtual} & 22.9& 23.8& 13.7& 25.6& 21.7& 8.5& 23.1& 29.8& 50.4
&26.7\\
    MixMatch~\cite{berthelot2019mixmatch} & 21.7& 25.9& 23.4& 23.1& 19.1& 35.7& 43.0& 42.5& 42.4  &34.4\\
    UDA~\cite{xie2020unsupervised} & 27.7& 32.8& 38.3& 41.3& 25.2& \underline{37.0}& 41.6& 43.3& 52.8
&40.0\\
    FixMatch~\cite{sohn2020fixmatch} & \underline{28.0}& \underline{35.5}& 38.8& 42.8& 27.8& 35.8& \underline{44.5}& \underline{44.2}& 56.0 
&\underline{41.6}\\
    FlexMatch~\cite{zhang2021flexmatch} & 25.6& 31.8& 37.8& \underline{44.5}& 25.3& 33.3& 42.6& 43.7& 56.8 
&41.6\\
    FreeMatch~\cite{wang2022freematch} & 26.7& 35.4& \underline{39.7}& 43.1& 25.8& 33.3& 42.1& 43.2& 53.6 
&40.7\\
    SoftMatch~\cite{chen2023softmatch} & 24.6& 28.3& 34.3& 34.6& 26.8& 32.4& 37.5 & 36.1  & 57.2
&36.4\\
 InfoMatch~\cite{han2024infomatch}& 24.4& 32.4& 35.3& 41.0 &22.8& 31.3& 36.7& 36.4 & 55.2
&41.5\\
 FineSSL~\cite{gan2024erasing}& 27.9& 30.9& 32.3& 33.3& \underline{28.1}& 31.1& 31.9& 33.1 & \underline{57.6}&32.4\\
    \midrule
    APLT (Ours) & \textbf{34.1}& \textbf{43.8}& \textbf{52.9}& \textbf{57.3}& \textbf{34.7}& \textbf{50.5}& \textbf{57.8}& \textbf{63.9} & \textbf{59.2}&\textbf{43.8}\\
    \bottomrule
  \end{tabular}
  \end{adjustbox}
\end{table*}

\subsection{Phase Iteration}

In our framework, the offline clustering phase and the online model training phase are performed alternately. Specifically, during model training, the memory-based classifier remains fixed and is used only to produce predictions. During the offline clustering, we both recalculate pseudo-labels based on the most recent model and update the prototypes of the memory-based classifier. Unlike FixMatch, this approach avoids creating pseudo-labels, updating the memory-based classifier, and producing predictions in the same iteration and manner, thereby mitigating overfitting to incorrect pseudo-labels. Furthermore, because deep networks possess a strong capacity to overfit, updating the prototypes and pseudo-labels too frequently can also lead to overfitting issue. Consequently, we run the online phase every 10 training epochs instead of every epoch. 

During testing, we discard the parametric classifier and use the outputs of the memory-based classifier as the final predictions.

\section{Experiments}

\subsection{Dataset and Experimental Setup}

\textbf{Benchmarks:} We build semi-supervised micro-action recognition datasets based on three datasets, \textit{i.e.}, MA-52~\cite{guo2024benchmarking}, iMiGUE~\cite{liu2021imigue}, and SMG~\cite{chen2023smg}. \textit{Note that, many classes in these three datasets are not satisfied for semi-supervised learning, as they do not have insufficient samples.} For example, in the SMG dataset, actions like \textit{Touching or covering} and \textit{Pulling shirt collar} have only 11 instances each. 
Hence, we only select classes with sufficient samples to form the datasets, results in MA-12, iMiGUE-11, and SMG-5, each of which has 12, 11, and 5 classes respectively. More details of are provided in the Appendix.

\noindent\textbf{Experimental Setup}. For each video, we use 8-frame clips for training and testing. We use the ResNet-18~\cite{he2016deep} as the backbone in default.
For the warm up stage, we train the model with 15 epochs with only the FixMatch loss function. We then introduce our asynchronous learning strategy and train the model with additional 40 epochs. The model is updated by SGD optimizer with a momentum of 0.9 and a weight decay of 0.0005. The learning rate is set to 0.002 and follows a cosine decay schedule. The offline clustering phase is implemented after every 10 training epochs.

\subsection{Comparison with State of The Arts}

\noindent \textbf{Comparison with Traditional SSL Methods.} We first compare our method with 11 state-of-the-art SSL approaches designed for image classification. For a fair comparison, we use the same backbone and experimental settings (e.g., data augmentation, optimizer, and training epochs) for all methods. The results on MA-12, SMG-5, and iMiGUE-11 are shown in Table~\ref{exp_MA-52}. Our method clearly outperforms all baselines in every setting. For instance, with only 10\% labeled data on MA-12, APLT surpasses FixMatch~\cite{sohn2020fixmatch} by 6.1\%. Moreover, most of the compared methods fail to improve on MA-12 with 10\% and 25\% labeled data, reflecting the challenges posed by SSMAR. In contrast, our method significantly boosts the baseline under both conditions. We further demonstrate that APLT benefits from a more powerful backbone. For example, using ResNet-50 instead of ResNet-18 yields an additional 6.6\% improvement on MA-12 with 50\% labeled data. However, most of the other methods show only marginal or even negative gains when switching to ResNet-50.

\noindent\textbf{Comparison with SSL Methods for Action Recognition.} We also compare with the art SSL methods designed for action recognition, including TCL~\cite{singh2021semi} , LTG~\cite{xiao2022learning} and SVFormer~\cite{xing2023svformer}. Results on MA-12 are shown in Table ~\ref{semi-supervised action recongnition methods.}. We can observe that our approach consistently achieves higher accuracies over all settings over the compared methods in which LTG and SVFormer use more complex backbones (3D-ResNet-18 or VIT-B). This further demonstrates the superiority of our method in SSMAR over previous SSL methods.

\begin{figure*}[!htb]
\centering
\includegraphics[width=1.0\linewidth]{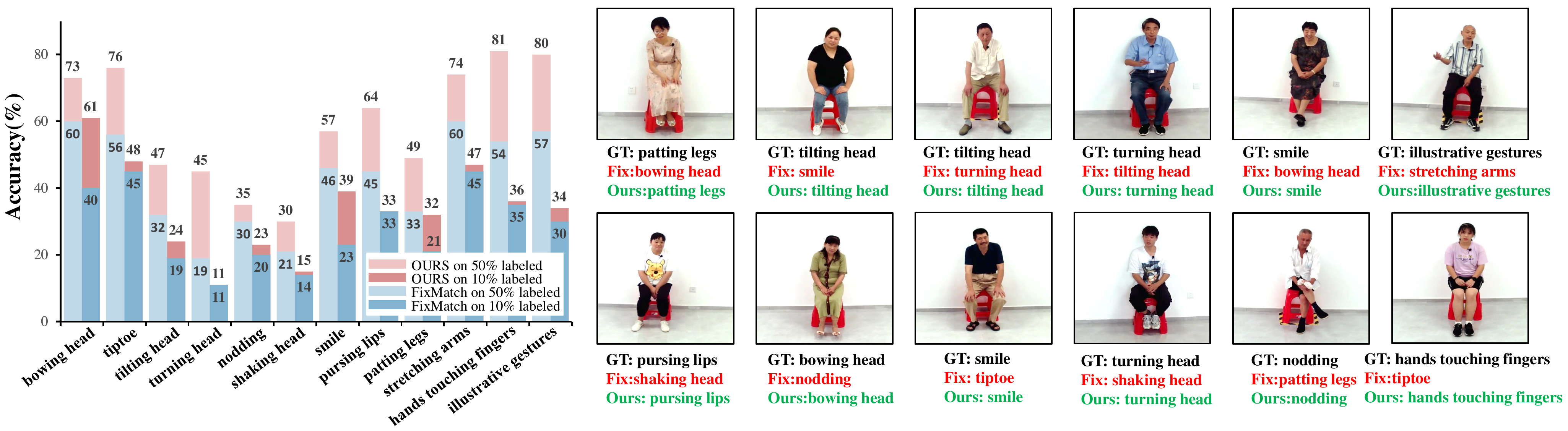}

\caption{\textbf{Left:} Class accuracy comparison between APLT with FixMatch for MA-12 with 10\% and 50\% labeled data. \textbf{Right:} Visualization of the predictions of APLT and FixMatch. The two methods are trained with ResNet-18.
}

\label{fig:Class_top1_change}
\end{figure*}

\noindent\textbf{Further Comparison with FixMatch}. Fig.~\ref{fig:Class_top1_change} (left) illustrates the class-specific gains of APLT compared to FixMatch~\cite{sohn2020fixmatch}. We can find that the majority of classes benefit from our method and  only two classes show comparable performance to FixMatch.
Fig.~\ref{fig:Class_top1_change} (right) provides a qualitative comparison between our APLT and FixMatch. Our approach accurately identifies different micro-actions that are misclassified by  FixMatch.

\begin{table}
  \caption{Comparisons with state-of-the-art SSL methods for action recognition methods on MA-12 ($\uparrow$).}
  
  \label{semi-supervised action recongnition methods.}
  \centering
   \begin{adjustbox}{width=0.48\textwidth}
  \begin{tabular}{lccccc}
    \toprule
    \multirow{2}{*}{\textbf{Method}} &  \multirow{2}{*}{\textbf{Backbone}} & \multicolumn{4}{c}{\textbf{Labeled Ratio}}\\
    \cmidrule(r){3-6}
  &     &\textbf{10\%}&\textbf{25\%}&   \textbf{40\% }&    \textbf{50\% }\\
   \midrule
 Baseline (Labeled Only)& ResNet-18& 24.8 & 30.7 & 35.8 &36.3 \\
  \midrule
   
    TCL\cite{singh2021semi}
&  ResNet-18& 16.3&25.0& 32.1& 37.3
\\
    
    LTG~\cite{xiao2022learning} 
& 3D-ResNet-18& 9.5&18.7&32.9&  34.9\\
 SVFormer~\cite{xing2023svformer} & ViT-B&  32.3&37.3& 38.3&44.9\\ 
    \midrule
    APLT (Ours) & ResNet-18& \textbf{34.1}&\textbf{43.8}&\textbf{52.9}&\textbf{57.3}\\
    \bottomrule
  \end{tabular}
  \end{adjustbox}
  
\end{table}

\subsection{Ablation Study}

\noindent \textbf{Analysis of  Components of Non-Parametric Clustering.} In Table~\ref{Experimental results on the MA-52}, we evaluate the effectiveness of the components of the proposed non-parametric clustering. Based on the FixMatch (+SSL), we further train the model with variants of our method, in which we generate pseudo-labels by different methods, including pure k-means (+KM), semi-supervised k-means (+SSKM), and SSKM with labeled-augmentation (+LA) and self-adaptive thresholding (+SAT). We also evaluate the effect of training the model with weak or strong augmentation. We make the following conclusions. First, semi-supervised learning obtains limited improvement for SSMAR. Second, using non-parametric clustering to generate pseudo-labels can significantly improve the accuracy. Third, updating the model with strong augmentations achieves slightly better results. Fourth, all the proposed components can consistently increase the performance and our full method obtains the best results on all settings.

\begin{table}
  \caption{Ablation study on MA-12 ($\uparrow$). \textbf{SSL:} FixMatch, \textbf{KM:} K-Means, \textbf{SSKM:}  Semi-Supervised K-Means, \textbf{W:} Weak augmentation, \textbf{S:} Strong augmentation, \textbf{LA:} Labeled-Augmentation, \textbf{SAT:} Self-Adaptive Thresholding.}
  \label{Experimental results on the MA-52}
  \centering
  \small
   \begin{adjustbox}{width=0.4\textwidth}
  \begin{tabular}{lccc}
    \toprule
    \multirow{2}{*}{\textbf{Method}} & \multicolumn{3}{c}{\textbf{Labeled Ratio}} \\
    \cmidrule(r){2-4}
  &    \textbf{25\%}&   \textbf{40\% }&    \textbf{50\% }\\
 
    \midrule
    \textbf{Baseline} (Labeled Only)    & 32.0& 36.0&41.4\\
     \midrule
    \textbf{SSL}&35.5 &38.8   &  42.8 \\
    \textbf{SSL} + \textbf{KM} &39.8&45.9    &50.1  \\
    \textbf{SSL} + \textbf{SSKM (W)}&40.2 &48.0   &51.7  \\
    \textbf{SSL} + \textbf{SSKM (S)}&41.0 &50.7 & 53.8 \\
    \textbf{SSL} + \textbf{SSKM (S)} + \textbf{LA}&42.5 &52.1 & 56.0 \\
    \textbf{SSL} + \textbf{SSKM (S)} + \textbf{SAT}&41.7&51.8  & 54.6 \\
    \textbf{SSL} + \textbf{SSKM (S)} + \textbf{LA} + \textbf{SAT}&\textbf{43.8}&\textbf{52.9}& \textbf{57.3}\\
    
    \bottomrule
  \end{tabular}
  \end{adjustbox}
  
\end{table}

\noindent\textbf{Effectiveness of Non-Parametric Classifier and Asynchronous Strategy.}  The proposed non-parametric classifier and the asynchronous pseudo-labeling model training strategy are both critical components of our approach. In Table~\ref{Effect of asynchronous update strategy on MA-12.}, we evaluate the contributions of these two techniques. Specifically, when applying the asynchronous strategy to FixMatch, we generate pseudo-labels via the parametric classifier in an offline manner and keep them fixed during training. Conversely, when using a synchronous strategy in our method, we take the non-parametric classifier’s online outputs as pseudo-labels, rather than the clustering results. We observe the following: 1) Using our non-parametric classifier substantially improves accuracy over FixMatch (Baseline), even under the synchronous strategy; 2) Both FixMatch and our APLT benefit from the asynchronous strategy, with APLT achieving a higher performance gain. These findings validate the effectiveness of both the proposed non-parametric classifier and the asynchronous strategy. Furthermore, when pseudo-labels and classifier outputs are generated through different mechanisms (\textit{e.g.}, our APLT), the asynchronous strategy delivers even stronger benefits.

\begin{table}[!t]
  \caption{Evaluation on synchronous strategy (SYN) and asynchronous strategy (ASY) on MA-12.}
  
  \label{Effect of asynchronous update strategy on MA-12.}
  \small
  \centering
     \begin{adjustbox}{width=0.46\textwidth}
  \begin{tabular}{lcc}
    \toprule
     \multicolumn{3}{c}{\textbf{MA-12 (50\% labeled data)}}\\
    \cmidrule(lr){1-3}   \textbf{Method}&     \textbf{Update Strategy}&   \textbf{Top-1 Acc.}\\
    \midrule
      FixMatch (Baseline)&	 SYN&42.8\\
 FixMatch & ASY&\textbf{43.4}\\

 \midrule
 APLT (Ours)&
 SYN&54.3\\
    APLT (Ours)& ASY&\textbf{57.3}\\
    \bottomrule
  \end{tabular}

  \end{adjustbox}
\end{table}

\noindent\textbf{Further Evaluation on Non-Parametric Classifier.} To further evaluate the benefit of the non-parametric classifier, we evaluate two variants. 
1) Variant I: We remove the non-parametric  classifier but further use the pseudo-labels generated by the clustering to train the parametric classifier that is used for testing. 
2) Variant II: We train the model with only the basic FixMatch loss, \textit{i.e.}, the model is trained without the loss of $\mathcal{L}^{margin}$. However,  we implement clustering with the final model and build the non-parametric classifier for testing. Results in Table~\ref{Effect of memory-based prototype classifier on MA-12.} show that without using the non-parametric classifier during training significantly reduces the accuracies. This indicates the importance of the non-parametric classifier in our approach. On the other hand, only using the non-parametric classifier can also achieve a clearly higher results than the baseline, further indicating the high-quality of the pseudo-labels generated by our non-parametric clustering.

\begin{table}
  \caption{Effect of non-parametric  classifier on MA-12. Variant I: Directly applying clustering labels on parametric classifier. Variant II: Only using non-parametric classifier during testing.}
  
  \label{Effect of memory-based prototype classifier on MA-12.}
  \centering
   \begin{adjustbox}{width=0.38\textwidth}
  \begin{tabular}{lccc}
    \toprule
    
    \multirow{2}{*}{\textbf{Method}} & \multicolumn{3}{c}{\textbf{Labeled Ratio}} \\
    \cmidrule(r){2-4}
  &    \textbf{25\%}&   \textbf{40\% }&    \textbf{50\% }\\
  \midrule
   Baseline & 30.7 & 35.8 & 36.3\\
   \midrule
    Variant I of APLT & 39.3& 44.0& 50.8 \\
    
    Variant II of APLT &37.6&44.9&  48.1\\ 
    \midrule
    APLT (Ours)&\textbf{43.8}&\textbf{52.9}&\textbf{57.3} \\
    \bottomrule
  \end{tabular}
  \end{adjustbox}
  
\end{table}

\section{Conclusion}

In this work, we introduce the Semi-Supervised MAR (SSMAR) setting. Through our evaluation, we show that traditional Semi-Supervised Learning (SSL) methods are prone to overfitting on inaccurate pseudo-labels. To overcome this, we present the Asynchronous Pseudo Labeling and Training (APLT) framework, which decouples the pseudo-labeling process from model training. In the offline pseudo-labeling phase, we propose a semi-supervised clustering approach to generate accurate pseudo-labels. During the online model training phase, we optimize the model with the proposed memory-based prototype classifier and the generated pseudo-labels. By alternating between pseudo-labeling and training phases asynchronously, APLT effectively mitigates error accumulation and enhances the accuracy. Extensive experiments on three MAR datasets validate the superiority of APLT, achieving significant performance gains over state-of-the-art SSL methods.


\bibliography{egbib}
\bibliographystyle{iclr2022_conference}

\end{document}